\documentclass[journal]{IEEEtran}
\usepackage{multirow}

\usepackage{amsmath,bm}
\usepackage{dsfont}
\usepackage{color} 
\usepackage[switch]{lineno}  %
\usepackage{graphicx}  
\usepackage{enumerate}
\usepackage{bbold}
\usepackage{array}
\makeatletter
\newcommand{\thickhline}{%
	\noalign {\ifnum 0=`}\fi \hrule height 1pt
	\futurelet \reserved@a \@xhline
}
\usepackage{amssymb}
\usepackage[pagebackref=true,breaklinks=true,letterpaper=true,colorlinks,bookmarks=false,citecolor=blue,linkcolor=blue]{hyperref}
\usepackage{cite}
\usepackage{enumitem}

\begin{document}
	
	\title{From Multi-View to Hollow-3D: \\ 
		Hallucinated Hollow-3D R-CNN for 3D \\
		Object Detection}
	
	\author{Jiajun Deng,~\IEEEmembership{Graduate Student Member,~IEEE},
		Wengang Zhou,~\IEEEmembership{Senior Member,~IEEE},\\
		Yanyong Zhang,~\IEEEmembership{Fellow,~IEEE},
		and~Houqiang Li,~\IEEEmembership{Fellow,~IEEE}
		\thanks{This work was supported in part by the National Key R\&D Program of China under Contract 2017YFB1002202, in part by the National Natural Science Foundation of China under Contract 61836011 and 62021001,  in part by the Youth Innovation Promotion Association CAS under Grant 2018497, and in part by the Anhui Provincial Development and Reform Commission 2020 New Energy Vehicle Industry Innovation Development Project ``Key System Research and Vehicle Development for Mass Production Oriented Highly Autonomous Driving''. It was also supported by the GPU cluster built by MCC Lab of Information Science and Technology Institution, USTC. (Corresponding authors: Wengang Zhou and Houqiang Li.)}
		\thanks{J. Deng, W. Zhou and H. Li are with the CAS Key Laboratory of Technology in Geo-Spatial Information Processing and Application System, Department of Electronic Engineering and Information Science, University of Science and Technology of China, Hefei 230026, China (e-mail: dengjj@mail.ustc.edu.cn; zhwg@ustc.edu.cn; lihq@ustc.edu.cn).}
		\thanks{Y. Zhang are with the Department of Computer Science,  University of Science and Technology of China, Hefei 230026, China (e-mail: yanyongz@ustc.edu.cn).}
		}
	
	\maketitle
	
	\begin{abstract} 
As an emerging data modal with precise distance sensing, LiDAR point clouds have been placed great expectations on 3D scene understanding. However, point clouds are always sparsely distributed in the 3D space, and with unstructured storage, which makes it difficult to represent them for effective 3D object detection. To this end, in this work, we regard point clouds as hollow-3D data and propose a new architecture, namely Hallucinated Hollow-3D R-CNN ($\text{H}^2$3D R-CNN), to address the problem of 3D object detection. In our approach, we first extract the multi-view features by sequentially projecting the point clouds into the perspective view and the bird-eye view. Then, we hallucinate the 3D representation by a novel bilaterally guided multi-view fusion block. Finally, the 3D objects are detected via a box refinement module with a novel Hierarchical Voxel RoI Pooling operation. The proposed  $\text{H}^2$3D R-CNN provides a new angle to take full advantage of complementary information in the perspective view and the bird-eye view with an efficient framework. We evaluate our approach on the public KITTI Dataset and Waymo Open Dataset. Extensive experiments demonstrate the superiority of our method over the state-of-the-art algorithms with respect to both effectiveness and efficiency. The code will be made available at \url{https://github.com/djiajunustc/H-23D_R-CNN}.

	\end{abstract}
	
	
	\IEEEpeerreviewmaketitle

	\section{Introduction}
	\IEEEPARstart{L}{iDAR} has become a vital component in the perception system of autonomous vehicle and intelligent robots. The point clouds generated by the laser scanner of a LiDAR can precisely measure the distance from the ego-sensor to the environment, thus they are applicable for 3D scene understanding, especially for 3D object detection. However, despite the advantage of capturing precise position information, point clouds data are notorious for their sparsity and unstructured storage, which leads to challenging feature representation learning. Besides, the real-time application poses the requirement of efficiency for 3D object detectors. 
	
	A series of approaches start with projecting 3D points into regular grids of either the bird-eye view~\cite{chen2017multi,simon2018complex,yang2018pixor,lang2019pointpillars,chu2021neighbor} or the perspective view~\cite{meyer2019lasernet}, and then exploit the similar architecture as 2D object detectors to generate predictions. Since they only rely on 2D CNNs for feature extraction, these methods are relatively efficient. However, the over-squeezed information in each single view usually leads to unsatisfied performance on distinguishing and localizing objects in 3D space~\cite{deng2020voxelrcnn,Shi_2020_CVPR}. 
	Avoiding the disadvantage of squeezed spatial information, another series of approaches extract 3D contextual features in 3D space. Specifically, they leverage 3D CNNs to process 3D voxels~\cite{zhou2018voxelnet,yan2018second,shi2020points,deng2020voxelrcnn}, or make use of the advanced point-based operators~\cite{qi2017pointnet,qi2017pointnet++,thomas2019kpconv,liu2020closer} to abstract the raw point clouds~\cite{shi2019pointrcnn,Yang_2019_ICCV,Shi_2020_CVPR,Yang_2020_CVPR}. The algorithms following this paradigm are generally more reliable thanks to the preserved 3D structure contexts. However, the runtime latency of 3D feature extraction is much higher than operating on 2D representations.
	
		\begin{figure}[t]
		\centering {\includegraphics[width=0.48\textwidth]{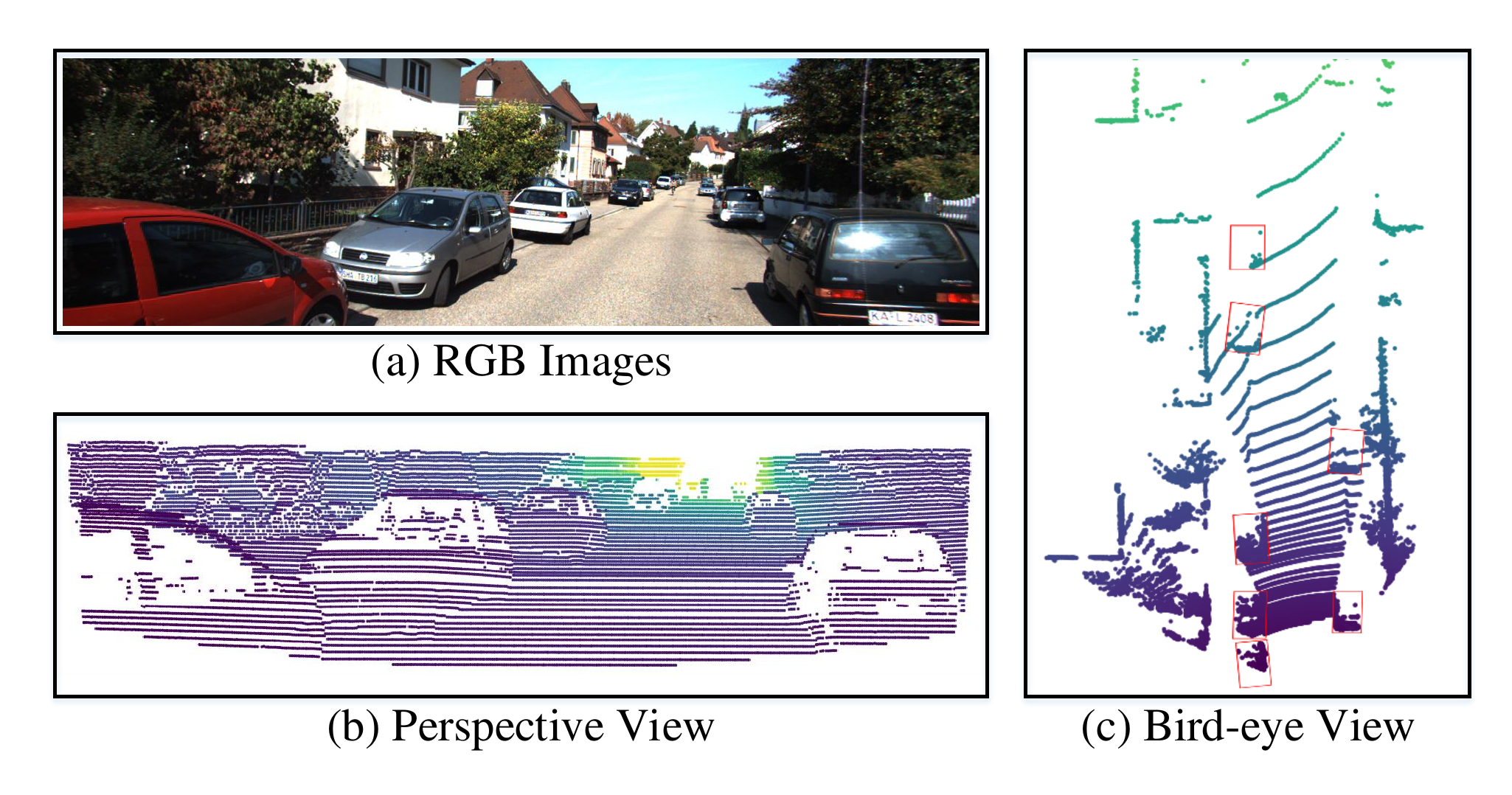}}
		\caption{An illustration of (a) the RGB image, (b) the perspective view of point clouds and (c) the bird-eye view of point clouds (better viewed in color). The red boxes in (c) indicate the locations of cars. }
		\label{fig:intro}
	\end{figure}

The dilemma leaves a valid problem to highlight both the accuracy and efficiency when designing a 3D object detector. To alleviate this, we revisit a long-term ignored property of LiDAR point clouds that the points are only reflected on the nearing surface of objects. We name this property as \emph{\textbf{hollow-3D property}}. Since the LiDAR point clouds are hollow-3D data, we can efficiently hallucinate the 3D representation by combining features from projected views, and meanwhile preserve the 3D spatial context. Formally, in this paper, the perspective view and the bird-eye view are explored. As shown in Figure~\ref{fig:intro}(b), the projected points are densely distributed in the perspective view, which looks similar to the RGB image in Figure~\ref{fig:intro}(a). Obviously, the features from the perspective view are rich in semantic context. On the contrary, as shown in Figure~\ref{fig:intro}(c), projected points in the bird-eye view are extremely sparse, making it difficult to distinguish the semantic components. Meanwhile, from this view, the size of objects are consistent regardless of their distance to the ego-sensor, and the objects are not likely to overlap with each other. Thus, features extracted from the bird-eye view are applicable to predict the scales and locations of objects~\cite{chen2020every,zhou2020end,Shi_2020_CVPR}.

By consolidating the idea of leveraging the hollow-3D property of point clouds and complementary information of multi-view projection, we introduce Hallucinated Hollow-3D R-CNN ($\text{H}^2$3D R-CNN) --- a new architecture that novelly hallucinates 3D representation from the features of perspective view and bird-eye view, and performs detection following the region-based paradigm~\cite{ren2015faster,dai2016r,cai2018cascade}. Specifically, $\text{H}^2$3D R-CNN begins with sequentially projecting the point clouds into the perspective view and bird-eye view for 2D feature extraction. In designing this module, we share the features between Region Proposal Network (RPN) and the backbone of bird-eye view, so that the cost of RPN is negligible. In particular, we pass the abstracted semantic information from the perspective view to the bird-eye view along with the raw point features, which facilitates the bird-eye-view feature extraction and region proposal generation. Then, the bilateral guided multi-view fusion block takes the obtained multi-view features as input, and capitalizes on cross-view bilateral guidance to hallucinate hollow-3D features. After multi-view fusion, our $\text{H}^2$3D R-CNN refines the region proposals to generate the final predictions. Distinctly, the RoI feature of each region proposal is extracted from the hollow-3D features with hierarchical voxel RoI pooling, a novel operator that takes hierarchical region proposal partition for further acceleration. The design of $\text{H}^2$3D-RCNN makes the best of Hollow 3D representation and strides a careful balance between accuracy and computation cost.
	
	In summary, we make three-fold contributions:
	 \begin{itemize}[noitemsep,nolistsep]
		\item  We novelly leverage the hollow-3D property of LiDAR point clouds to develop a new 3D object detection framework, namely  $\text{H}^2$3D R-CNN.
		\item We present an elegant view of how to take full advantage of complementary information from the perspective view and bird-eye view, which is a non-trivial problem not yet fully understood.
		\item We conduct extensive experiments on both KITTI Dataset and Waymo Open Dataset. The encouraging average precision, together with the low latency for processing, verify the superiority of our proposed hallucinated hollow-3D feature representation. 
	\end{itemize}

	\section{Related Work}

	In the literature, 3D object detectors applied on LiDAR point clouds are commonly categorized according to their data representation formats, \emph{i.e.}, point-based and voxel-based. Differently, since we focus on hallucinating the hollow-3D representation with multi-view features, we categorize the related approaches by the space for feature extraction and the space for making prediction. Specifically, the bird-eye view, multiple views and 3D space are involved.
	
	\begin{figure*}[t]
		\centering {\includegraphics[width=1.00\textwidth]{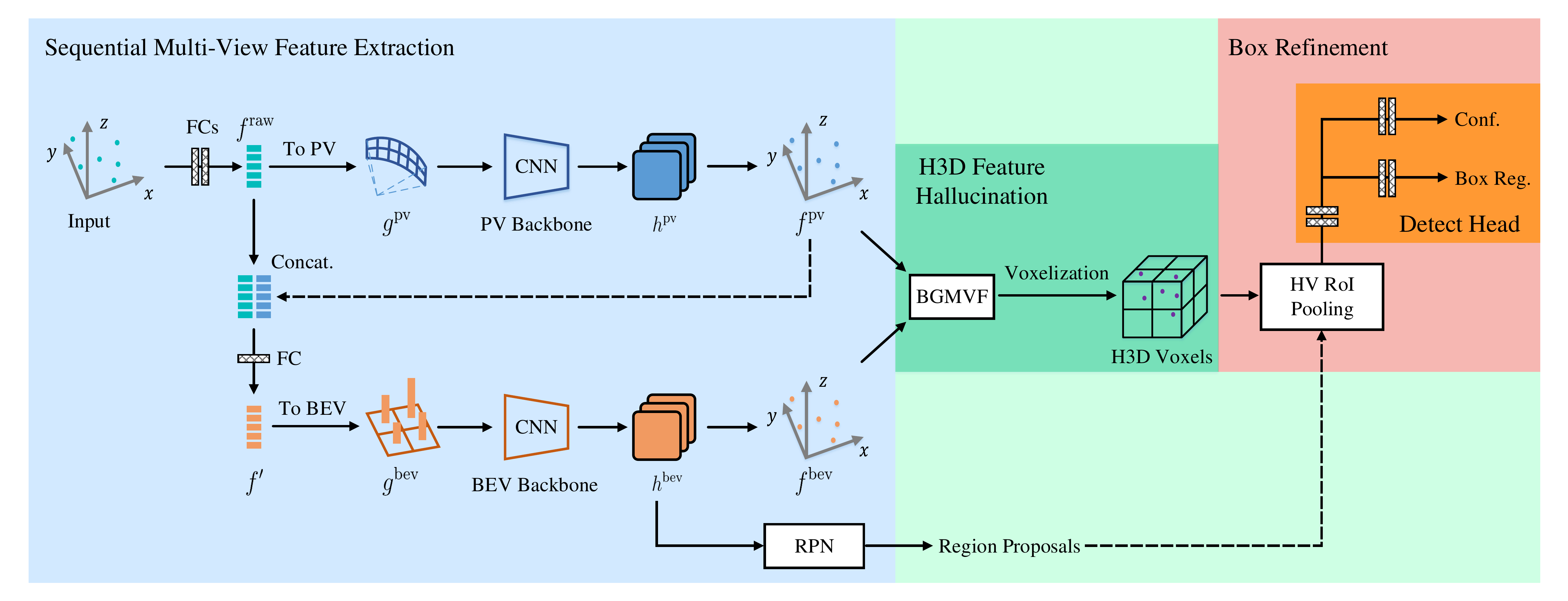}}
		\caption{An overview of  the proposed $\text{H}^2$3D R-CNN framework (better viewed in colors). Our framework includes three main components: (a) a sequential multi-view feature extraction module, (b) a H3D feature hallucination module, and (c) a box refinement module. }
		\label{fig:framework}
		\vspace{-0.1in}
	\end{figure*}
	
	\subsection{Feature Extraction from The Bird-Eye View, and Making Final Prediction from The Bird-Eye View}
	The approaches following this paradigm are characterized by projecting point clouds into the bird-eye view, and then operating on the bird-eye-view feature representation.
	The early work MV3D \cite{chen2017multi} projects the point clouds to 2D bird-eye-view grids and places lots of predefined 3D anchors for generating 3D bounding boxes. PIXOR~\cite{yang2018pixor} encodes each voxel grid as occupancy and Complex YOLO~\cite{simon2018complex} converts point clouds into bird-eye-view RGB maps. In contract to previous methods with hand-crafted bird-eye-view representation, PointPillars \cite{lang2019pointpillars} groups points as ``pillars'', and extracts the pillar feature with a simplified PointNet module~\cite{qi2017pointnet,qi2017pointnet++}.  The recent work HVNet~\cite{Ye_2020_CVPR} extends PointPillars by fusing pillars of different scales at point-wise level to generate pseudo images in multiple resolutions.
	
	\subsection{Feature Extraction in 3D Space, and Making Final Prediction from The Bird-Eye View}
	Different from leveraging bird-eye-view representation for both feature extraction and making final prediction, the methods following this paradigm first abstract features volumes in 3D space, and then squeeze the 3D features into 2D BEV features to perform detection. VoxelNet~\cite{zhou2018voxelnet} exploits 3D voxel partition to divide unordered points into regular grids and uses a tiny PointNet to abstract points in each voxel as a compact feature representation. 
	Several 3D convolutional middle layers are applied before converting the 3D features into the bird-eye view for 3D feature extraction. 
	However, only a small fraction of voxels are non-empty in large 3D space, which makes applying 3D CNN inefficient. To tackle this problem, SECOND~\cite{yan2018second}  replaces the conventional 3D convolutional layers with sparse convolution operation. SA-SSD~\cite{He_2020_CVPR} further introduces an auxiliary loss at the training stage to facilitate preserving 3D structure information.
	
	\subsection{Feature Extraction in 3D Space, and Making Final Prediction in 3D Space}
	Besides feature extraction, predicting/refining boxes in the 3D space is also proven to be of great significance for precise 3D detections. 
	The pioneer work  F-PointNet \cite{qi2018frustum} extrudes each 2D region proposal from the RGB image to a 3D viewing frustum, and utilizes PointNet \cite{qi2017pointnet} for 3D bounding box prediction. 
	To get rid of 2D region proposals from RGB images, PointRCNN \cite{shi2019pointrcnn}  introduces a 3D region proposal network based on PointNet++ \cite{qi2017pointnet++}, and devises point cloud RoI pooling to extract 3D region features for each proposal. 
	The following work STD \cite{Yang_2019_ICCV} proposes a sparse-to-dense strategy for better proposal refinement. 3DSSD~\cite{Yang_2020_CVPR} is the first one-stage 3D object detector operating on the raw points, and it introduces F-FPS as a complement of D-FPS~\cite{qi2017pointnet++} to facilitate key points sampling.
	PV-RCNN~\cite{Shi_2020_CVPR} on the one hand abstracts the 3D voxel features into a set of sparsely sampled keypoints, and on the other hand generates region proposals with bird-eye-view voxel features. The region-wise features are extracted by grid RoI pooling from the keypoints to make further refinement over the preliminary proposals. Voxel R-CNN~\cite{deng2020voxelrcnn} finds that the coarse granularity of voxels is sufficient for high performance 3D object detection, and introduces a novel Voxel RoI Pooling operation to accelerate the process of extracting region-wise features from 3D feature volumes.
	
	\subsection{Feature Extraction from Multiple Views, and Making Final Prediction from The Bird-Eye View}
	The most related works to ours are MVF~\cite{zhou2020end} and Pillar-od~\cite{wang2020pillar}.
	These two methods both leverage the multi-view features to augment the raw point features before converting the points into bird-eye-view pillars representation~\cite{lang2019pointpillars}. After that, MVF performs one-stage object detection with anchor, while Pillar-od follows the anchor-free paradigm~\cite{tian2019fcos}.
	
	 In contrast to MVF and Pillar-od that leverage multi-view information for ameliorating detection capability of bird-eye-view features, the proposed $\text{H}^2$3D R-CNN hallucinates 3D representation by considering features of perspective view and bird-eye view with respect to properties of each view, and efficiently refine the candidate predictions in the 3D space. Therefore, our approach can be categorized as ``\textbf{\emph{Feature Extraction from Multiple Views, and Making Final Prediction in 3D Space}}''. A detailed comparison between MVF and our method is discussed in~\ref{sec:compare_to_mvf}.

	\section{Our Approach}
	
	\subsection{Overview}\label{sec:overview}
	Motivated by the fact that LiDAR point clouds are only reflected on the nearing surface of objects (\emph{i.e.}, hollow-3D property), we leverage the 2D features from the bird-eye view and the perspective view to hallucinate the hollow-3D representation. The resulting hollow-3D representation is applicable in scene understanding with LiDAR point clouds, and in this paper we focus on region-based 3D object detection. 
	
	As shown in Figure~\ref{fig:framework}, the proposed $\text{H}^2$3D R-CNN framework includes three main components: (a) a sequential multi-view feature extraction module, (b) a hollow-3D (H3D) feature hallucination module, and (c) a box refinement module. Given the input point clouds, we sequentially project the points into the perspective view and the bird-eye view, and leverage 2D backbone networks for feature extraction. Particularly, the point-wise features projected into the bird-eye view not only comes from the raw points, but also includes the perspective-view features. Thus the abstracted semantic information from the perspective view is passed to the bird-eye view along with the position information from the raw points. Besides, we share the features between bird-eye-view feature extraction and the Region Proposal Network (RPN), so that the cost of RPN is negligible. After obtaining the multi-view 2D features,  Bilaterally Guided Multi-View Fusion (BGMVF) is exploited to hallucinate hollow-3D features of each point. The hollow-3D features are then voxelized into hollow-3D voxels. At the end, the box refinement module capitalizes on Hierarchical Voxel RoI Pooling (HV RoI Pooling) to extract proposal features from the hollow-3D voxels, and makes further box refinement in the detect head. Distinctly, HV RoI Pooling divides each region proposal into grids of hierarchical sizes (\emph{i.e.}, a coarse partition and a fine partition). The coarse partition exploits a large grouping radius, while the fine partition exploits a small grouping radius. The final RoI feature is a sophisticated combination of the hierarchical grids.

	In the following, we first describe the preliminary in subsection~\ref{sec:pre}. Then, we elaborate the detailed design of sequential multi-view feature extraction, H3D feature hallucination and box refinement in subsections~\ref{sec:extraction}, \ref{sec:fusion} and~\ref{sec:head}, respectively. Finally, we present the training objective of our method in subsection~\ref{sec:loss}.

	\subsection{Preliminary}\label{sec:pre}

	\subsubsection{Multi-view Grid Indexing}
\begin{figure}[t]
		\centering {\includegraphics[width=0.45\textwidth]{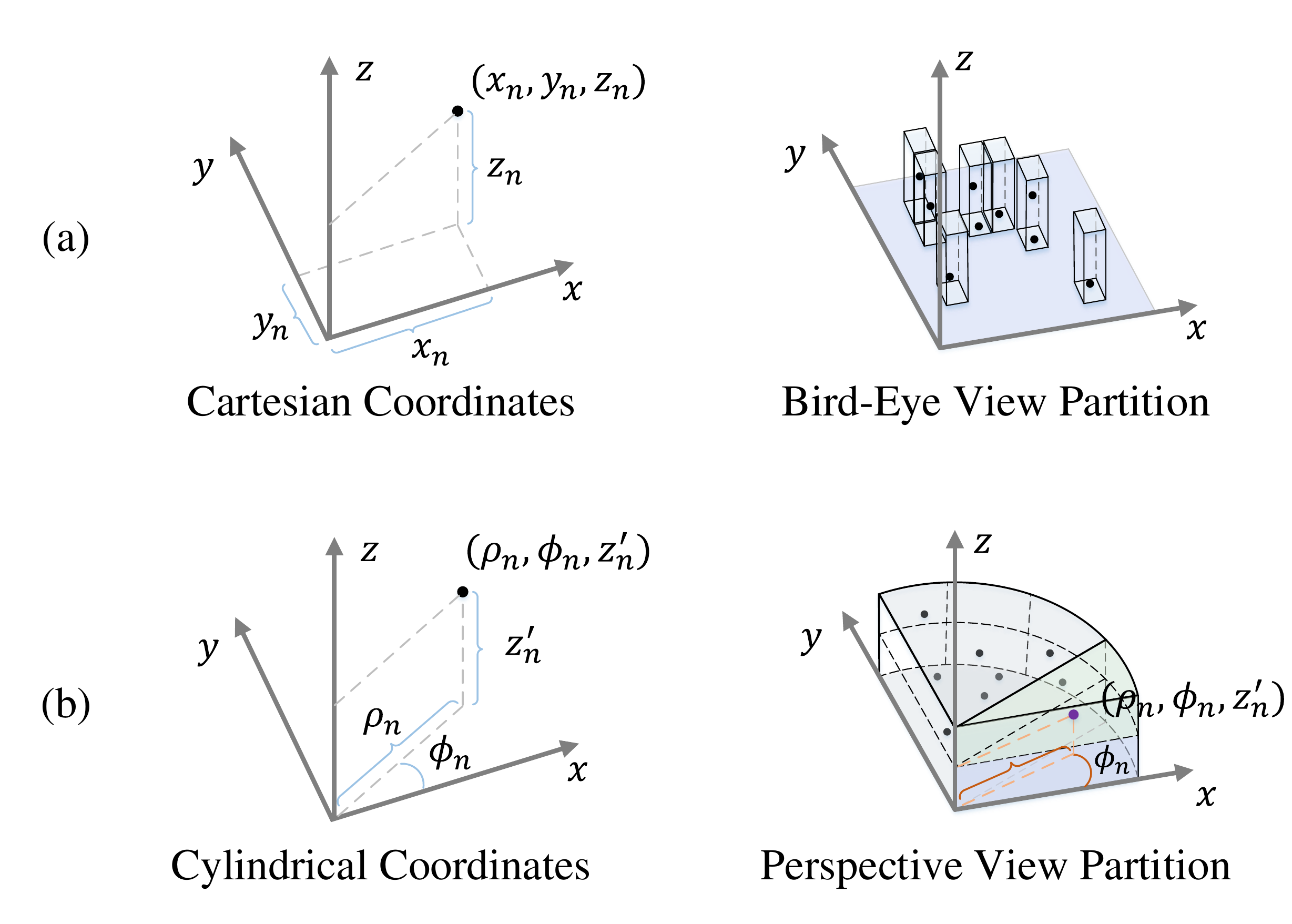}}
		\caption{An illustration of coordinates systems and partition methods in different views. (a) In the bird-eye view, we use Cartesian Coordinates and project the points into the $X-Y$ plane. (b) In the perspective view, we exploits Cylindrical Coordinates, and the two axes for the cylindrical surface is azimuth $\Phi$ and height $Z$, respectively.
		}
		\label{fig:coordinates}
	\end{figure}

	To project the 3D points into the perspective view and the bird-eye view, we first introduce how to index them. As illustrated in Figure~\ref{fig:coordinates}, we exploit Cartesian coordinates system in bird-eye view, and leverage cylindrical coordinates system in perspective view. Given a point $\bm{p}_n$ with coordinates of $(x_n, y_n, z_n)$ in the Cartesian coordinate system, its coordinates in the cylindrical coordinates system is computed as follows:
	\begin{equation}
		\left\{
		\begin{aligned}
			\rho_n & = \sqrt{x_n^2+y_n^2}, \\
			\phi_n & = \text{arctan}{\frac{y_n}{x_n}}, \\
			z'_n & =  z_n,
		\end{aligned}
		\right.			
	\end{equation}
	where $\rho_n$ is the range of the point, and $\phi_n$ is the azimuth ($\Phi$ axis) on horizontal plane. Let us denote the point cloud range of the X and Y axes as $[X_\text{min},X_\text{max}]$ and $[Y_\text{min},Y_\text{max}]$, and denote the range of $\Phi$ and Z axes as $[\Phi_\text{min},\Phi_\text{max}]$ and  $[Z_\text{min},Z_\text{max}]$. The corresponding projected indices of $\bm{p}_n$ in each view are computed as:
	\begin{equation}
		\left\{
		\begin{aligned}
			l_n & = \frac{\phi_n - \Phi_\text{min}}{V^{\phi}}, \\
			k_n & =\frac{z'_n - Z_\text{min}}{V^{z}},
		\end{aligned}
		\right.		
		and
		\hspace{1em}
		\left\{
		\begin{aligned}
			i_n & = \frac{x_n - X_\text{min}}{V^x}, \\
			j_n & = \frac{y_n - Y_\text{min}}{V^y},
		\end{aligned}
		\right.		
	\end{equation}
   where $(V^{\phi},V^{z})$ and $(V^x,V^y)$ are the input voxel size for the perspective view partition and the bird-eye view partition, $(l_n, k_n)$ is the projected 2D index of the perspective view, and $(i_n, j_n)$ is the projected 2D index of the bird-eye view. Accordingly, the quantized grid indices are $(\lfloor l_n \rfloor , \lfloor k_n \rfloor)$ and  $(\lfloor i_n \rfloor , \lfloor j_n \rfloor)$, respectively. 
	The grid indices are leveraged for voxelization, and the projected indices are used for mapping 2D features back to points. Here we exploits bilinear interpolation to map the 2D feature maps back to each point.
	
	\subsubsection{Unique-Scatter Voxelization}
	To enable dynamic voxelization~\cite{zhou2020end,wang2020pillar,zhang2020polarnet} and gradient back-propagation from 2D feature maps to each point, we exploit the unique-scatter paradigm for voxelization. Let us take the perspective view as an example. Given the perspective-view grid indices $\{(\lfloor l_n \rfloor , \lfloor k_n \rfloor)\}$ and point-wise features $\{\bm{f}_n\}$ of point clouds, we first remove the duplicate grid indices and obtain a set of unique ones. After that, for each unique grid, we perform max pooling over the points within this grid, which is also known as a ``Scatter-Max'' operation. The grid feature $\bm{g}(l^*,k^*)$ of the $(l^*,k^*)$-th cell is computed as:
	\begin{equation}\label{equ:voxelization}
		\bm{g}(l^*,k^*) = max(\{ \bm{f}_n ~|~ (\lfloor l_n \rfloor,\lfloor k_n \rfloor)=(l^*,k^*) \}),
	\end{equation}
	where $max(\cdot)$ means max pooling. The bird-eye view voxelization and 3D voxelization are computed following the same paradigm, but with different indices and features.

	\subsection{Sequential Multi-view Feature Extraction}\label{sec:extraction}
	
	\subsubsection{Perspective View}
	The introduced sequential multi-view feature extraction starts with the perspective view. Given the input points $\{\bm{p}_n\}$, an MLP is first applied to each point to generate raw point features $\{\bm{f}^\text{raw}_n\}$. The MLP is composed of two fully connected layers, each of which followed by a batch normalization (BN) layer and a ReLU activation layer. Note that  we omit the BN and ReLU layers in Figure~\ref{fig:framework}. Once obtaining the raw point features $\{\bm{f}^\text{raw}_n\}$, we exploit the voxelization paradigm as in Equation~\eqref{equ:voxelization} to project the 3D points into the perspective view, and generate the input feature map $\bm{g}^\text{pv}$. Particularly, for the cells in $\bm{g}^\text{pv}$ that without projected points, we fill it with zeros. After that, a 2D  backbone network takes $\bm{g}^\text{pv}$ as input to extract the perspective-view ( (PV)) features. We use bilinear interpolation to generate point-wise PV features $\{\bm{f}^\text{pv}_n\}$ according to the projected index $\{(l_n, k_n)\}$.
	
	As shown in Figure \ref{fig:intro}(b), the points are densely distributed in the perspective view, and the dense observation eases the difficulty to distinguish the foreground and background regions. However, the perspective view suffers from occlusion problem and the size of objects changes with the distance to the LiDAR sensor. Considering these advantages and disadvantages, the perspective-view feature is not applicable to directly generate box predictions, but benefits abstracting semantic information from the point clouds, which is significantly difficult in the sparse 3D space or the bird-eye view. Therefore, we put the perspective view at the beginning of the multi-view feature extraction module, so that the extracted perspective-view features can not only play an important role in hallucinating hollow-3D features, but also be leveraged to augment the inputs of the bird-eye view. Specifically, given the raw point features $\bm{f}^\text{raw}$ and perspective-view features $\bm{f}^\text{pv}$ of each point, we concatenate them, and apply a fully-connected layer followed by BN and ReLU layers to generate the input of the bird-eye view. This process is computed as follows:
		\begin{equation}\label{eq:bev_input}
		\bm{f}'= \text{ReLU}(\text{BN}(\text{FC}([\bm{f}^\text{raw},\bm{f}^\text{pv}]))),
	\end{equation}
where $\bm{f}'$ is the point-wise input feature of the bird-eye view, $[\cdot,\cdot]$ indicates concatenation operation.\newline

	\subsubsection{Bird-Eye View} The bird-eye-view (BEV) feature extraction is conducted following the perspective view. Given the point features $\{\bm{f}'_n\}$ obtained as described in Equation~\eqref{eq:bev_input}, we perform voxelization to project these points into the BEV feature maps $\bm{g}^\text{bev}$ according to their BEV grid indices $\{(\lfloor i_n \rfloor,\lfloor j_n \rfloor)\}$. Then, a 2D backbone network is applied on $\bm{g}^\text{bev}$ to extract BEV features. Similar to the perspective view, we also utilize the bilinear interpolation to get the point-wise BEV features $\{\bm{f}^\text{bev}_n\}$, and the corresponding project indices are $\{(i_n,j_n)\}$.
	
	In the bird-eye view, the sizes of objects are consistent regardless of their distance to the LiDAR sensor, and they are not like to overlap with each other. Therefore, the bird-eye view is especially applicable for generating region proposals. In particular, as shown in the bird-eye view branch of Figure \ref{fig:framework}, we build the Region Proposal Network (RPN) on top of the BEV backbone network, so that the RPN shares the feature with 2D BEV feature extraction, which results in negligible cost for region proposal generation.
	
	Typically, the design of our sequential multi-view feature extraction module follows the art of compactness. We make the most of both views to play extra roles in our $\text{H}^2$3D R-CNN. Specifically, the perspective view features are leveraged to augment the bird-eye view features, and the bird-eye view features are shared with the RPN to generate region proposals. Consequently, such a design enables our method to stride a careful balance between accuracy and computation cost.

	\subsection{H3D Feature Hallucination}\label{sec:fusion}
	After sequentially extracting the PV feature $\bm{f}^{pv}$ and BEV feature $\bm{f}^{bev}$,  we feed both of them into the H3D feature hallucination module to fuse their complementary information and generate H3D features. Generally, there are various choices to simply fuse the multi-view feature, \emph{i.e.}, concatenation, element-wise summation and element-wise multiplication. However, the PV features are semantic-intensive and with height information of each instance, while the BEV features are sensitive to the location of objects in the horizontal plane. Adopting simple fusion operations mentioned above will inevitably ignore the differences between these two views, incurring hard optimization and sub-optimal performance.

	\subsubsection{BGMVF} 
\begin{figure}[t]
	\centering {\includegraphics[width=0.38\textwidth]{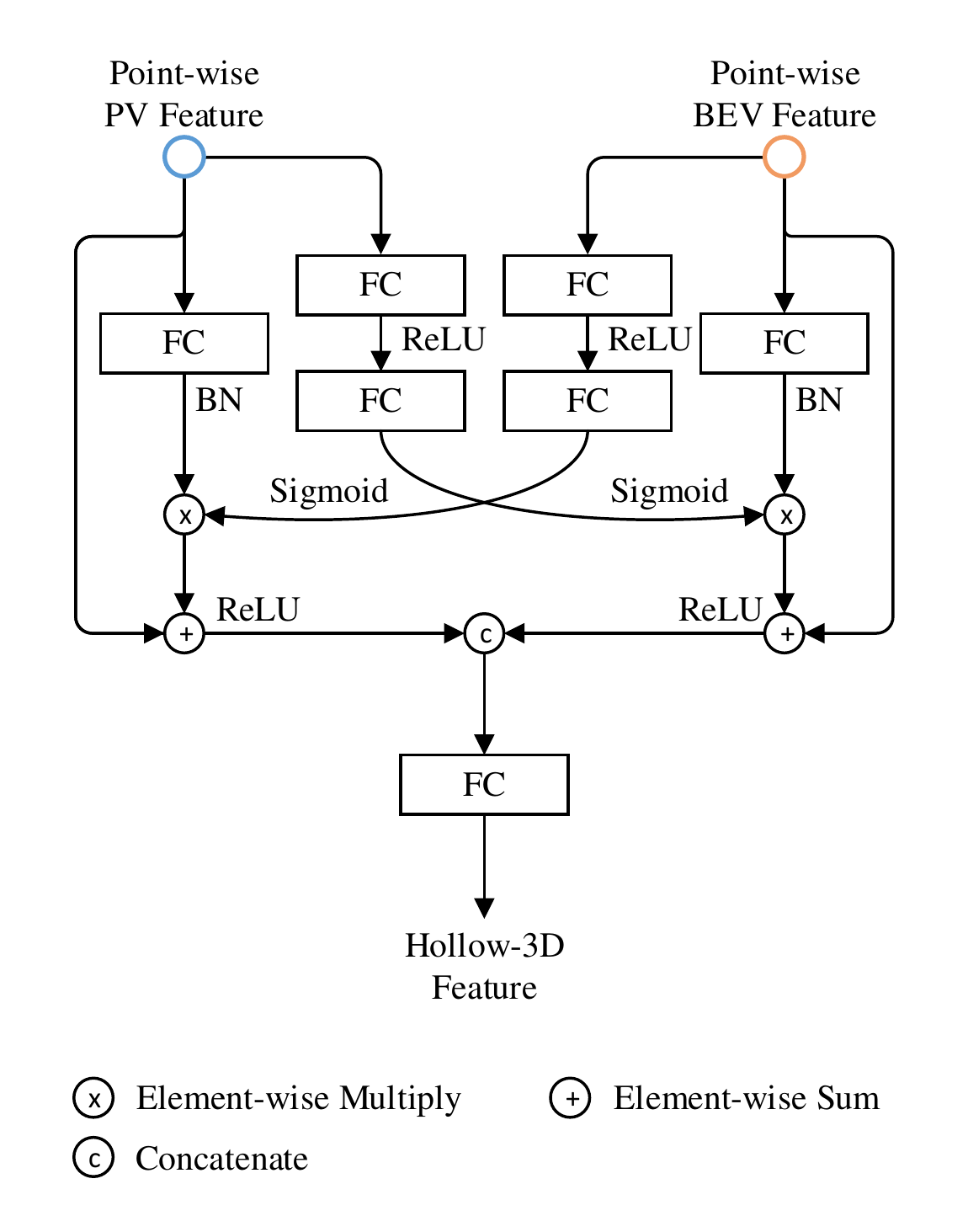}}
	\caption{An illustration of the Bilaterally  Guided Multi-View Fusion module for hollow-3D feature hallucination. Notation: \textit{FC} means the fully connected layer, \textit{BN} denotes the batch normalization, \textit{ReLU} is the ReLU activation function, and \textit{Sigmoid} indicates the Sigmoid activation function.
	}
	\label{fig:bgmvf}
\end{figure}

	Motivated by the architecture in~\cite{anderson2018bottom,lee2018stacked,chen2020expressing} to fuse the multi-modal features with cross-modal attention, we devise the Bilaterally Guided Multi-View Fusion (BGMVF) block to combine the multi-view features. The detailed design of BGMVF block is shown in Figure~\ref{fig:bgmvf}. In BGMVF, we utilize the cross-view feature guidance to comprise a light weight gating mechanism over channels for feature selection. The BGMVF layer benefits the efficient interaction between these two views, and improves the representation capability of hollow-3D (H3D) features. A similar cross-stream gating operation is also proposed in SAOA~\cite{wang2020symbiotic}. However, SAOA leverages the cross-stream gating operation to further calibrate the object-centric features, while our BGMVF is devised to introduce the information of one view to guide underlining the complementary information of the other view.
	
	\subsubsection{H3D Voxelization} 
	The output of BGMVF layer is the point-wise H3D features. We perform voxelization again according to the 3D grid indices $\{(\lfloor i_n \rfloor , \lfloor j_n \rfloor , \lfloor k_n \rfloor)\}$ to generate H3D voxels. Here the 3D index $(\lfloor i_n \rfloor , \lfloor j_n \rfloor , \lfloor k_n \rfloor)$ is a hybrid of the BEV index $(\lfloor i_n \rfloor , \lfloor j_n \rfloor)$ and the index $\lfloor k_n \rfloor$ from the perspective view. We use the voxelized H3D features for further operation instead of the point-wise H3D features, since the voxels are more robust to the anormal nodes and more efficient for proposal feature extraction~\cite{deng2020voxelrcnn}. 

	\subsection{Box Refinement}\label{sec:head}
	Since the output of H3D feature hallucination module is the voxelized H3D features, one straightforward way is to apply Voxel RoI Pooling~\cite{deng2020voxelrcnn} for RoI feature extraction. In the conventional Voxel RoI Pooling, each region proposal is first divided into $G$$\times$$G$$\times$$G$ regular grids, and the center point of each grid is taken as the grid point. Then, the voxel query is performed by each grid point to group neighbor voxels within a given query radius (distance threshold). The grouped voxel features are aggregated to the corresponding grid points, and the grid features together construct the RoI feature of each region proposal. In~\cite{deng2020voxelrcnn}, multi-scale query ranges (\emph{i.e.}, a small query range and a large one) are adopted to enable flexible receptive fields.\
	However, the non-empty voxels within a small and a large query range centered on the same grid point can be highly reduplicate, which may incur the significance of multi-scale grouping. Besides, the computational cost of grouping neighbor voxels grows \textit{cubically} with the partition grid number $G$ and the query range $r$. When both $G$ and $r$ are large, lots of time is wasted on grouping neighbor voxels instead of actual feature extraction.\newline
	
	\begin{figure}[t]
		\centering {\includegraphics[width=0.49\textwidth]{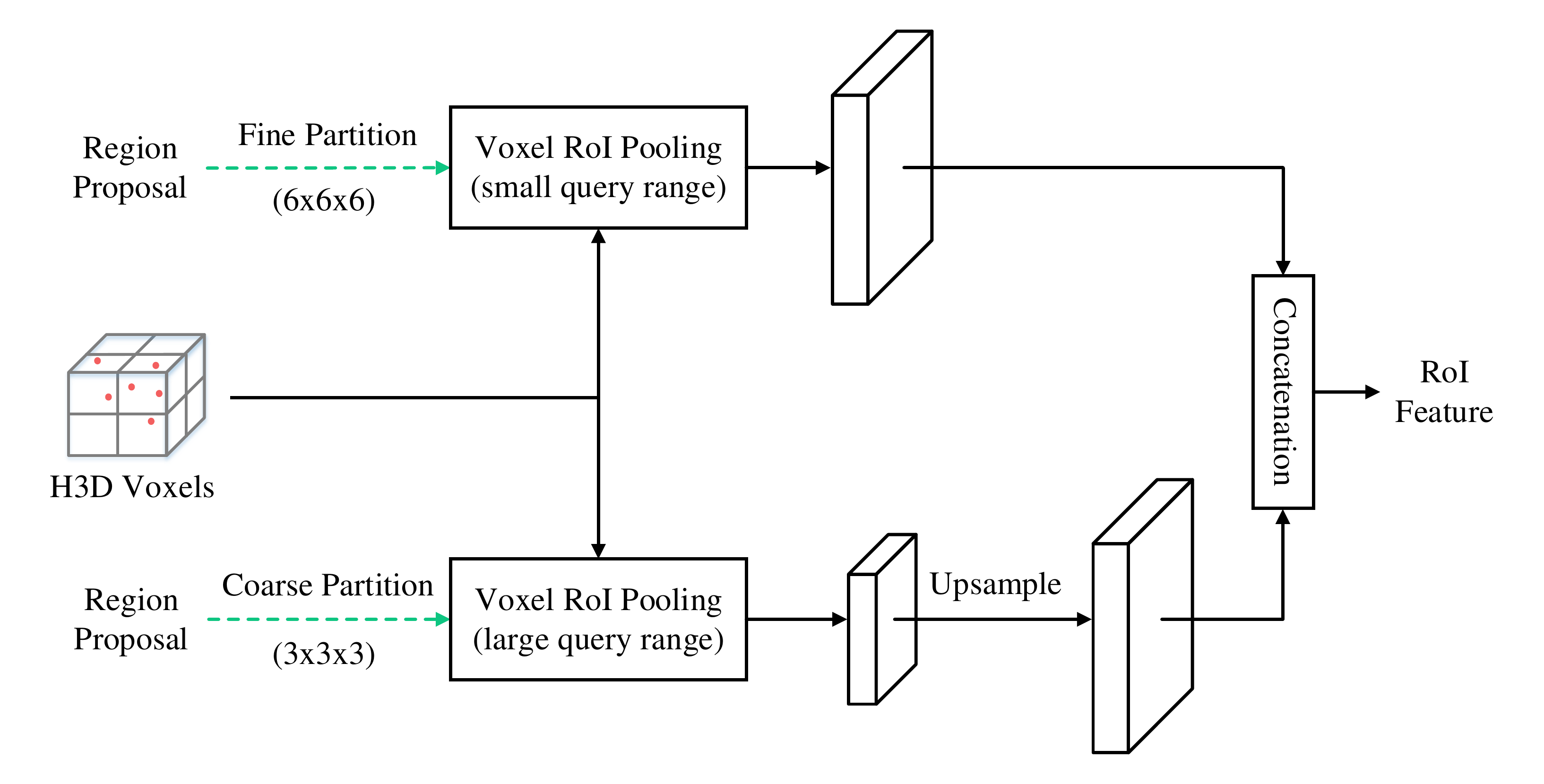}}
		\caption{An illustration of Hierarchical Voxel RoI Pooling (HV RoI Pooling).
		}
		\label{fig:hvroi_pooling}
	\end{figure}

	\subsubsection{HV RoI Pooling} Based on our analysis on the bottlenecks of conventional Voxel RoI Pooling, we devise Hierarchical Voxel RoI Pooling (HV RoI Pooling) to further reduce the computational cost of our box refinement module. As shown in Figure \ref{fig:hvroi_pooling}, the workflow of HV RoI Pooling consists of three steps: (1) we first divide the region proposal into the a coarse partition grids (\emph{i.e.}, 3$\times$3$\times$3) and a fine partition grids (\emph{i.e.}, 6$\times$6$\times$6); (2) we exploit a large query range to perform Voxel RoI Pooling on the coarse grids, and exploit a small range for the fine grids; (3) the features abstracted by the coarse partition are upsampled to the same scale as that of the fine one, and we concatenate them to construct the RoI features. Typically, our HV RoI Pooling on the one hand maintains the flexible receptive fields of multi-scale grouping, and on the other hand avoids the computational bottleneck of grouping voxels in a large range with fine grids partition.\newline 
	
	\subsubsection{Detect Head} After extracting the RoI features, the detect head predicts an IoU-related confidence score and performs further box coordinates regression for each region proposal. Our detect head adopts the same architecture as that in ~\cite{deng2020voxelrcnn,Shi_2020_CVPR} with several fully connected layers.

\subsection{Training Objective}\label{sec:loss}

	At the training stage, the proposed $\text{H}^2$3D R-CNN framework is end-to-end optimized, and the total training objective is computed as:
	\begin{equation}\label{loss_total}\small
		L_\text{Total} = L_\text{RPN} + L_\text{Head},
	\end{equation}
    where $L_\text{RPN}$ is the training objective of RPN~\cite{yan2018second,lang2019pointpillars,He_2020_CVPR}, and $L_\text{Head}$ is the training objective of the detect head~\cite{Shi_2020_CVPR,deng2020voxelrcnn}. We elaborate  $L_\text{RPN}$  and $L_\text{Head}$ as follows.

	\subsubsection{Objective of RPN} 
	The prediction of RPN involves classification and box regression for anchors. We leverage Focal Loss~\cite{lin2017focal} ($L_\text{focal}$) for classification, and exploit Smooth L1 Loss ($L_\text{loc}$) for the box regression. Specifically, the objective of RPN is computed as follows:
	\begin{equation}\label{equ:rpnloss}\small
		\begin{split}
			L_\text{RPN} =  \frac{1}{N_a}[ & \sum_i L_\text{focal}(\alpha_i, \alpha_i^*)\\
			&+  \mathbb{1}(\alpha_i^*\geq1)\sum_{i} L_\text{loc}(\beta_i,\beta_i^*)],
		\end{split}
	\end{equation}
	where $N_a$ is the number of foreground anchors, $\alpha_{i}$ and $\beta_{i}$ are the classification and regression predictions of the the $i$-th anchor box, $\alpha_{i}^*$ and $\beta_{i}^*$ are the corresponding classification label and regression target, and $\mathbb{1}(\alpha_i^*\geq1)$ indicates the regression loss is only calculated with the foreground anchors.
	
	\subsubsection{Objective of Detect Head} In the objective function of our detect head, the target confidence value of the $i$-th region proposal is first assigned according to its overlap ($\text{IoU}_i$) with the corresponding ground-truth box:
	\begin{equation}\label{lossofdetecthead}
		\gamma_i^*(\text{IoU}_i)=\left\{
		\begin{array}{lll}
			0 & & {\text{IoU}_i < \theta_l},\\
			\frac{\text{IoU}_i - \theta_l}{\theta_h-\theta_l} & & {\theta_l \leq \text{IoU}_i < \theta_h},\\
			1 & & {\text{IoU}_i > \theta_h},
		\end{array} \right.
	\end{equation}
	where $\theta_h$ and $\theta_l$ are the foreground and background IoU thresholds, respectively. Here we exploit Binary Cross Entropy Loss ($L_\text{bce}$) for confidence prediction, and use Smooth L1 Loss ($L_\text{loc}$) for box regression. Let us denote the confidence prediction as $\gamma_i$, and denote the regression prediction as $\upsilon_i$, the objective of our detect head is computed as:
	\begin{equation}\label{losshead}\small
		\begin{split}
			L_\text{head} = \frac{1}{N_s}[&\sum_{i} L_{\text{bce}}(\gamma_i,\gamma_i^*(\text{IoU}_i))\\
			&+\mathbb{1}(\text{IoU}_i\ge{\theta_{reg}})\sum_{i} L_{\text{reg}}(\upsilon_i, \upsilon_i^*)],
		\end{split}
	\end{equation}
	where $N_s$ is the number of sampled region proposals, and $\upsilon_i^*$ represents the regression target.

\section{Discussion}\label{sec:discussion}
In this section, we discuss about: (1) the differences between MVF~\cite{zhou2020end} and our proposed $\text{H}^2$3D R-CNN, and (2) the potential to generalize $\text{H}^2$3D R-CNN into point cloud sequences.

\subsection{Comparison with MVF}\label{sec:compare_to_mvf}
\begin{figure}[!t]
	\centering {\includegraphics[width=0.50\textwidth]{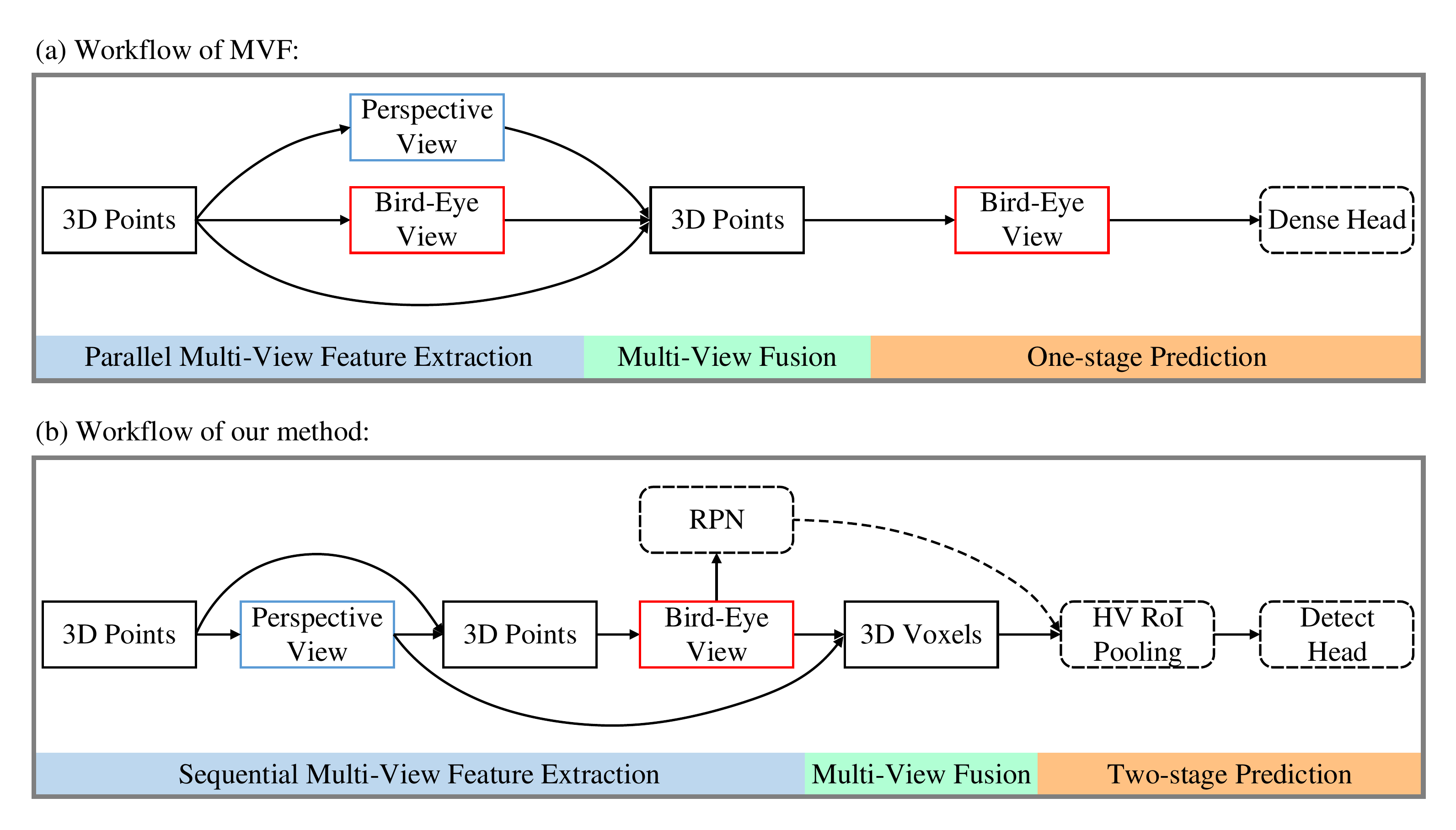}}
	\caption{An illustration of workflows of (a) MVF and (b) our $\text{H}^2$3D R-CNN.
	}
	\label{fig:workflow}
\end{figure}

The workflows of MVF and our method have been illustrated in Figure~\ref{fig:workflow}. The main differences between MVF and our $\text{H}^2$3D R-CNN can be concluded into three folds:

\textbf{Motivation.} MVF intends to improve the PointPillars, which directly projects 3D raw points into the bird-eye view to make one-stage prediction, by concatenating the raw points features with the multi-view features for augmentation. Our $\text{H}^2$3D R-CNN is designed to take full advantage of the complementary property of the perspective view (semantic intensive) and the bird-eye view (size of objects are consistent regardless their distance to the ego-sensor) to generate a powerful 3D representation from multi-view features, which can get rid of the costing point-based operations and 3D convolutional networks.

\textbf{Architecture.} MVF exploits multi-view feature extraction in parallel.  The bird-eye view projection is performed twice in MVF (the first time for multi-view feature extraction and the second time for making one-stage prediction), which is redundant. In the contrast, our $\text{H}^2$3D R-CNN devises sequentially multi-view feature extraction. We make the most of both views to play extra roles besides constructing the Hollow-3D features. Specifically, the perspective-view features are concatenated with raw point features to enhance the bird-eye-view features, and the bird-eye-view features are additionally exploited for region proposal generation. Both of these two extra roles takes negligible cost, which makes our model compact and efficient.

\textbf{Paradigm.} MVF is a one-stage 3D object detector that performs detection by sliding windows (the dense head) on the bird-eye-view representation. Our $\text{H}^2$3D R-CNN follows the two-stage object detection paradigm. $\text{H}^2$3D R-CNN first generates region proposals from the bird-eye view, and then extracts the region-wise features from the voxelized hollow-3d features through HV RoI pooling. Finally, the features of each proposal is leveraged to make further box refinement.

\subsection{Potential of Generalizing into Point Clouds Sequences}
Exploiting temporal information is a promising way to further improve the 3D object detection in autonomous driving scenarios~\cite{huang2020lstm,yuan2020temporal,qi2021offboard,yang20213d}. Since there are three different formats of feature representation in $\text{H}^2$3D R-CNN, \emph{i.e.}, the single-view 2D feature maps, the voxelized H3D features and the proposal features, our method is of great potential and flexibility to generalize into applications of point cloud sequences.  Specifically, for the single-view 2D feature maps, the pixel-level feature aggregation by spatio-temporal sampling~\cite{bertasius2018object} or attention-based pixel alignment~\cite{guo2019progressive} are applicable. For the voxelized H3D features, the non-empty voxels can be represented by their center coordinates and the corresponding H3D features. The Point Spatio-Temporal Convolution~\cite{fan2012pstnet} can be applied on the non-empty H3D voxels from a sequence of adjacent frames to abstract the temporal coherence. In addition, the proposal features can be augmented by leveraging the cross-frame object relations~\cite{deng2019relation} for feature augmentation.

	\section{Experiments}
	
	\subsection{Datasets and Evaluation Metrics}
	
	\subsubsection{KITTI Dataset}
	The KITTI Dataset~\cite{geigerwe} contains both samples of RGB images and point clouds in urban driving scenes, and we only use the point clouds data in experiments. In this dataset, objects of 3 common categories in the street, \emph{i.e.}, car, pedestrian and cyclist, are annotated. The LiDAR point cloud data is officially split into 7481 training samples and 7518 testing samples. We follow the common setting~\cite{Shi_2020_CVPR,wang2020pillar,Yang_2019_ICCV,simon2018complex} to further divide the official training samples into a \textit{train} set with $3712$ samples and a \textit{val} set with $3769$ samples. In our analytical experiments, our model is trained on the \textit{train} set and evaluated on the \textit{val} set. For test server submission, we randomly select $80\%$ samples from the official training set to train our model and leave the others for model selection. We report the average precision of our method on both the official testing set and commonly used \textit{val} set for comparison with other state-of-the-art methods.

	\subsubsection{Waymo Open Dataset} \hskip -5pt
	The large-scale Waymo Open Dataset \cite{sun2020scalability} is recently released for promoting autonomous driving to the industrial production level. Different from KITTI Dataset that only provides annotations in camera FOV ($<90^\circ$), Waymo Open Dataset provides annotations for objects in the full $360^\circ$ panoramic view. The annotated point clouds data of this dataset are from $1,000$ sequences, consisting of $798$ sequences ($\sim$$158\text{k}$ point clouds samples) for training and 202 sequences ($\sim$$40\text{k}$ point clouds samples) for validation. The objects on the Waymo Open Dataset are split into two levels based on the number of points of a single object, where the LEVEL\_1 objects have more than 5 points while the LEVEL\_2 objects have 1$\sim$5 points. Since the amount of samples in this dataset is too large, we uniformly sample $\sim$$32\text{k}$ frames by an interval of 5 during training, and we evaluate our model on the whole validation set.

	\subsection{Implementation Details}\label{sec:implement}

\begin{table*}[]
	\centering
	
	\renewcommand\arraystretch{1.1}
	
	\caption{Performance comparison on the KITTI test set in terms of average precision (AP) calculated with recall 40 positions. The results are evaluated with the online KITTI test server.}
	\footnotesize
	\scalebox{0.97}[0.97]{
		\setlength\tabcolsep{2.9pt}
		\begin{tabular}{ l | c  c  c | c  c  c | c c c | c c c | c  c  c | c  c  c  }
			\hline
			\multirow{2}{*}{Method}			& \multicolumn{3}{c|}{$\text{Car-AP}_\text{3D}$ (\%) }		&\multicolumn{3}{c|}{$\text{Car-AP}_\text{BEV}$ (\%) }		&
			\multicolumn{3}{c|}{$\text{Ped.-AP}_\text{3D}$ (\%) }	& 
			\multicolumn{3}{c|}{$\text{Ped.-AP}_\text{BEV}$ (\%) } & \multicolumn{3}{c|}{$\text{Cyc.-AP}_\text{3D}$ (\%) }	& 
			\multicolumn{3}{c}{$\text{Cyc.-AP}_\text{BEV}$ (\%) } 
			 \\ 
				& Easy		& Mod.		& Hard		& Easy		& Mod.		& Hard		& Easy		& Mod.		& Hard	& Easy		& Mod.		& Hard & Easy		& Mod.		& Hard & Easy		& Mod.		& Hard\\ 	
			
			\hline \hline
			\textbf{Multi-modal Feature:} &  &  &  &  &  & &  &  &  &  &  & &  &  &  &  &  &\\
			MV3D \cite{chen2017multi} & 74.97 & 63.63 & 54.00 	&  86.62		&  78.93 			&  69.80	&  - 		&  - 		&  -	&  - 		&  - 		&  - 	&  - 		&  - 		&  -	&  - 		&  - 		&  - \\ 
			AVOD-FPN~\cite{ku2018joint}			&  83.07 & 71.76 & 65.73 		& 90.99		& 84.82			& 79.62	    & 50.46 & 42.27 & 39.04 & 58.49 & 50.32&  46.98     &   63.76	 	& 50.55		& 44.93	&  69.39	 	& 57.12		& 51.09	\\ 
			
			F-PointNet \cite{qi2018frustum}	    &   82.19   & 69.79   & 60.59		  & 91.17		& 84.67		& 74.77		& 50.53 & 42.15 & 38.08 & 57.13 & 49.57 & 45.48 &   72.27	& 56.12 & 49.01	& 77.26	& 61.37 & 53.78	\\ 
			
			F-ConvNet~\cite{wang2019frustum} & 87.36 & 76.39 & 66.69 & 91.51 & 85.84 & 76.11   &52.16 & 43.38 & 38.80 & 57.04 & 48.96 & 44.33 & 81.98 & 65.07 & 56.54& 84.16 & 68.88 & 60.05 \\
			
			UberATG-MMF \cite{liang2019multi} & 88.40 & 77.43 & 70.22	&  93.67		&  88.21 			&  81.99		&  - 		&  - 		&  - &  - 		&  - 		&  -  	&  - 		&  - 		&  - &  - 		&  - 		&  - \\
			
			\hline
			\textbf{3D Feature:} &  &  &  &  &  & &  &  & &  &  & &  &  & &  &  &\\

			SECOND-V1.5 \cite{yan2018second}   & 84.65 & 75.96 & 68.71 &  91.81 & 86.37 & 81.04 		&  - 		&  - 		&  - &  - 		&  - 		&  - &  - 		&  - 		&  - &  - 		&  - 		&  -\\
			Part-$A^2$ \cite{shi2020points}& 87.81 & 78.49 & 73.51 & 91.70 & 87.79 & 84.61 &53.10 & 43.35 & 40.06 & 59.04 & 49.81 &45.92  & 79.17 & 63.52 & 56.93 & 83.43 & 68.73 & 61.85 \\
			SA-SSD~\cite{He_2020_CVPR}  & 88.75 & 79.79 & 74.16 & 95.03 & 91.03 & 85.96 		&  - 		&  - 		&  - &  - 		&  - 		&  - &  - 		&  - 		&  - &  - 		&  - 		&  -\\
			
			3DSSD~\cite{Yang_2020_CVPR}  &88.36 & 79.57 & 74.55 &  -		&  - 			&  -		&  - 		&  - 		&  - &  - 		&  - 		&  -  &  - 		&  - 		&  - &  - 		&  - 		&  -\\
			PointRCNN~\cite{shi2019pointrcnn}  & 86.96 & 75.64 & 70.70 & 92.13 & 87.39 & 82.72 & 47.98 & 39.37 & 36.01 & 54.77& 46.13& 42.84 &  74.96 & 58.82 & 52.53 & 82.56 & 67.24 & 60.28 \\
			STD~\cite{Yang_2019_ICCV}  & 87.95 & 79.71 & 75.09 & 94.74& 89.19 & 86.42 & 53.29 & 42.47 & 38.35 & 60.02 & 48.74 & 44.55 & 78.69  &61.59 & 55.30 & 81.36 & 67.23 & 59.35 \\
			PV-RCNN~\cite{Shi_2020_CVPR} & 90.25 & 81.43 & 76.82 & 94.98 & 90.65 & 86.14  & 52.17 & 43.29 & 40.29 & 59.86 & 50.57 & 46.74 & 78.60 &63.71 &57.65 & 82.49 & 68.89 & 62.41 \\
			Point-GNN~\cite{Shi_2020_CVPR_PointGNN}  & 88.33 & 79.47 & 72.29 & 93.11 & 89.17 & 83.9 &51.92 & 43.77&40.14 & 55.36& 47.07 & 44.61 & 78.60 & 63.48 & 57.08 & 81.17 & 67.28 & 59.67 \\
			\hline
			\textbf{BEV Feature:} &  &  &  &  &  & &  &  &  &  &  & &  &  &  &  &  &\\
			PointPillars~\cite{lang2019pointpillars}				& 82.58 & 74.31 & 68.99 & 90.07 & 86.56 & 82.81 &51.45 & 41.92 & 38.89 & 57.60& 48.64 & 45.78  & 77.10 & 58.65 & 51.92	& 79.90 & 62.73 & 55.58	\\ 
			
			TA-Net~\cite{liu2020tanet}   & 83.81 & 75.38 & 67.66 & - & - & - & 54.92 & 46.67 & 42.42 & -&  -&- &73.84 & 59.86 &53.46 & - & - & -\\
            \hline
			\textbf{Multi-view Feature:}&  &  &  &  &  & &  &  &  &  &  & &&&&&&\\
			$\text{H}^2$3D R-CNN (ours)  & 90.43 & 81.55 & 77.22  &  92.85 & 88.87  & 86.07 & 52.75 & 45.26 & 41.56 & 58.14 & 50.43& 46.72 & 78.67 & 62.74 & 55.78 & 82.76 & 67.90 & 60.49   \\
			\hline
			
		\end{tabular}
	}
	
	\label{tab:kitti_test_3d}
\end{table*}

	\subsubsection{Voxelization}
	For the KITTI Dataset, the range of point clouds in Cartesian coordinates is first clipped into $[0.0, 70.4]m$ for the $X$ axis, $[-40.0,40.0]m$ for the $Y$ axis and $[-3.0,1.0]m$ for Z axis, and the range of azimuth in Cylindrical coordinates is $[-90,90]^\circ$ correspondingly. We set the voxel size to $(0.20m, 0.20m)$ for the bird-eye view ($X$ and $Y$ axes) and $(0.33^\circ, 0.10m)$ for the perspective view ($\Phi$ and $Z$ axes). For the Waymo Open Dataset, the range of point clouds in the Cartesian coordinates is first clipped into $[-75.2, 75.2]m$ for the $X$ and $Y$ axes, and $[-2, 4]m$ for the $Z$ axis. Since the Waymo Open Dataset provides box annotations for the $360^\circ$ panoramic view, we don't need to clip the azimuth range in the Cylindrical coordinates. The size of each voxel is set as the same as that in the KITTI Dataset, \emph{i.e.}, $(0.20m, 0.20m)$ in the bird-eye view and $(0.33^\circ, 0.10m)$ in the perspective view.
	
	\subsubsection{Augmented Input}
	The raw input representation of each point is $(x, y, z, r, t)$, where $(x, y, z)$ is the coordinates in the Cartesian coordinate system, $r$ is the reflection intensity and $t$ is the time stamp (no $t$ for the KITTI Dataset). We augment the raw input with $\phi, x^{bev}_d, y^{bev}_d, z^{pv}_d, and \phi^{pv}_d$, where $\phi$ is the azimuth in Cylindrical coordinate system, and $d$ subscript denotes the distance to the center of the belonging grid from the bird-eye view or the perspective view. In summary, the augmented input of our $\text{H}^2$3D R-CNN is $(x, y, z, \phi,  x^{bev}_d, y^{bev}_d, z^{pv}_d, \phi^{pv}_d, r, t)$.

		\begin{figure}
	\centering {\includegraphics[width=0.48\textwidth]{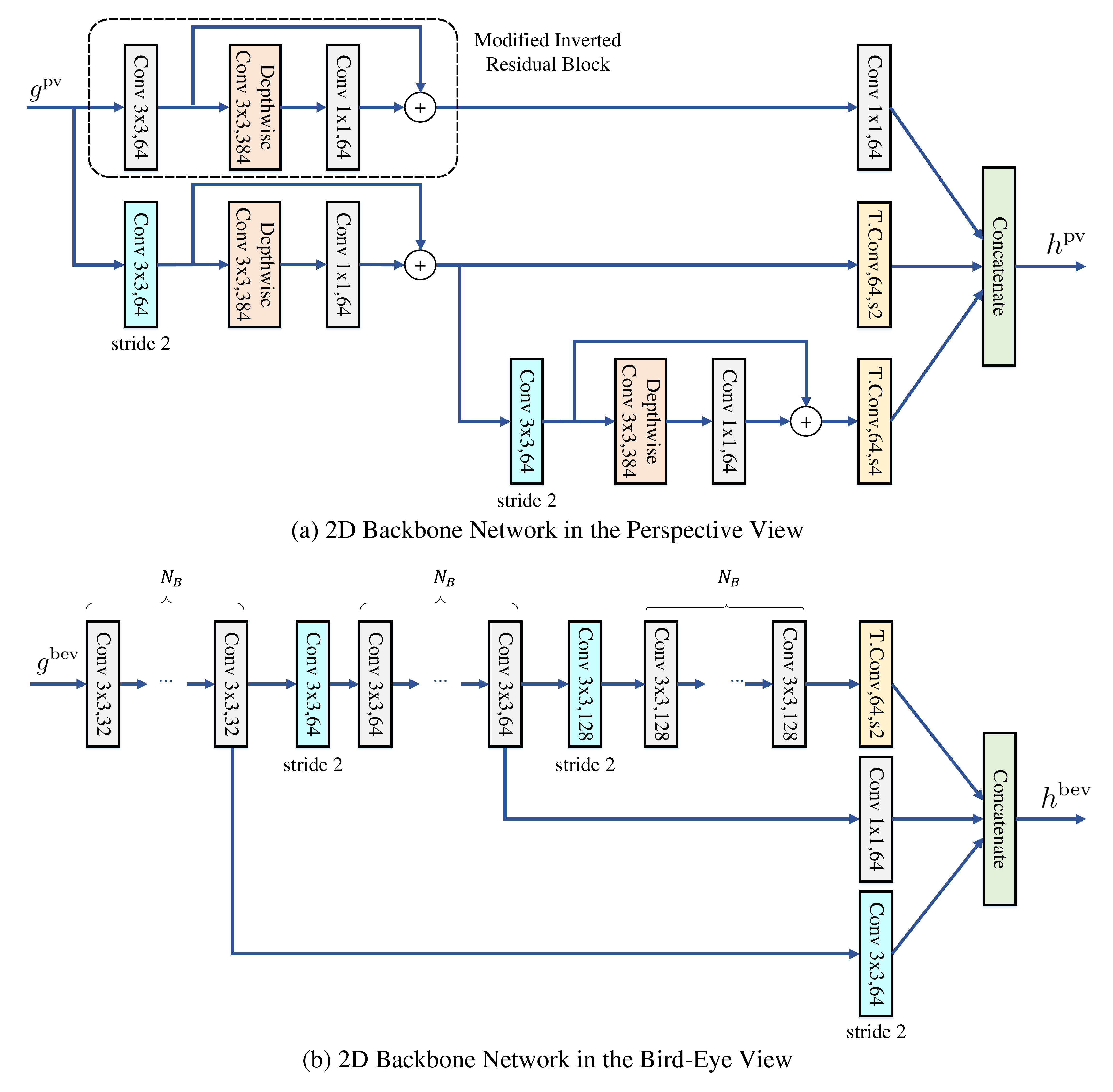}}
	\caption{An illustration of (a) 2D Backbone Network in the Perspective View and (b) 2D Backbone Network in the Bird-eye View .  
	}
	\label{fig:backbone_net}
\end{figure}

	\subsubsection{Network Architecture}
	The architecture of 2D backbone in the perspective view and the bird-eye view are shown in Figure \ref{fig:backbone_net}. The meta architecture of our 2D backbone networks adopts the design in~\cite{zhou2020end} and~\cite{lang2019pointpillars}. In the perspective view, as motivated by~\cite{sandler2018mobilenetv2}, we devise a modified inverted residual block (shown in the dash box) to increase the capacity of our backbone while keep the low latency. In the bird-eye-view network, there are three stages. The first stage keeps the same resolution along $X$ and $Y$ axes as the input, while the second and the third stages are half and 
	quarter the resolution of the first one, respectively. The layer number $N_B$ is set as 4, and the output dimension of convolution layers in these three stages are set as 32, 64 and 128. Each convolutional layer is followed by a batch normalization layer and a ReLU activation layer, except for the $1\times1$ convolutional layers in the modified inverted residual block (no ReLU activation layer here). We use the same architecture in both KITTI and Waymo Open Dataset. The network morphing and scaling may lead to further improvements in our $\text{H}^2$3D R-CNN, but it is not the focus in this paper. We leave it to the future investigation. The coarse grid partition in HV RoI Pooling is set as 3$\times$3$\times$3, and the fine one is set as 6$\times$6$\times$6. The query radius is set as 6 for the coarse partition and 3 for the fine partition. We randomly sample no more than 16 neighbor voxels for each queried grid point when performing HV RoI Pooling.
	

	\subsubsection{Training}
	The optimization algorithm is one-cycle Adam with initial learning rate 0.01. In the detect head, the foreground IoU theshold $\theta_h$, the background IoU theshold $\theta_l$, and the box regression IoU theshold $\theta_{reg}$ are set as 0.75, 0.25, 0.55, respectively. To update the detect head, we randomly sample 128 RoIs as the training samples. Within the sampled RoIs, 50\% are positive samples which have IoU$>$ $\theta_{reg}$ with the corresponding ground truth boxes. Following the widely adopted strategies in~\cite{lang2019pointpillars,He_2020_CVPR,Shi_2020_CVPR,Yang_2020_CVPR,deng2020voxelrcnn}, we conduct data augmentation in the training stage. For KITTI Dataset, the number of training epochs is 80 and the batch size is 32. For Waymo Open Dataset, the number of training epochs is 50 and the batch size is 24.

	\subsubsection{Inference}
	In the RPN, we adopt Fast NMS~\cite{bolya2019yolact}  with IoU threshold $0.7$ to generate no more than 128 region proposals as the inputs of our box refinement module. After performing box refinement, the conventional NMS is used with IoU threshold $0.1$ to remove the redundant predictions.

	\subsection{Comparison with State-of-the-art Methods}

   \subsubsection{KITTI Dataset. }
	We follow the common practice to report the average pricision on both the official \textit{test} set and the widely adopted \textit{val} set for comparision.  According to the official protocol, the IoU threshold is set as 0.7 for ``Car'', and the IoU threshold is set as 0.5 for ``Pedestrian'' and ``Cyclist''.
	
	Table~\ref{tab:kitti_test_3d} shows the 3D average precision (AP)$\footnote{The AP evaluation of the test server changes from recall 11 positions to 40 positions on 08/10/2019. To make fair comparison with the methods published before that day, we exploit the AP of recall 11 positions on the val set. }$ and BEV AP of state-of-the art methods on the official \textit{test} set. 
	In this table, we divide the methods into four groups according to the features leveraged in these models. Typically, the methods with 3D features consistently outperform that with only bird-eye-view features. Although our method only extracts features with 2D CNN from the perspective view and the bird-eye view, it can outperform most of the methods with 3D features. This result significantly validates the effectiveness of our proposed hollow-3D features hallucinated from the multi-view representation.
	 Remarkably, our method achieves state-of-the-art performance \emph{i.e.}, 90.43\%, 81.55\%, 77.22\% AP on easy, moderate and hard difficulty levels,  for the most important 3D object detection benchmark of the ``Car'' class. Meanwhile, our $\text{H}^2$3D R-CNN can run at a real-time speed (the runtime comparison is shown in Table~\ref{tab:val_r11}), \emph{i.e.}, 37.1 FPS with the NVIDIA RTX 2080 Ti GPU, which is 4.2 times faster than the strongest competitor PV-RCNN~\cite{Shi_2020_CVPR} (\emph{i.e.}, 8.9 FPS on the same device) and comparable to the fastest competitor PointPillars~\cite{lang2019pointpillars}.

	 Compared to ``Car'' class, the superiority of our approach is not that significant for ``Pedestrian'' and ``Cyclist'' classes. This is mainly because the scale of pedestrians and cyclists are much smaller than cars. When these two kinds of objects are distant to the ego-sensor, there are only few points reflected by them. Although the projected points in the perspective view are densely distributed, few points cannot provide strong semantic evidence for detecting these objects. To address this issue, a potential solution is to involve the RGB images to ameliorate the detection  precision of pedestrians and cyclists.
	
	As we mentioned in the related work, the most related methods to ours are MVF~\cite{zhou2020end} and Pillar-od~\cite{wang2020pillar}, which also exploit multi-view features. Thus, the proposed $\text{H}^2$3D R-CNN is supposed to be compared with these two methods. However, the results of these two approaches have not been submitted to the online test server, so that their detection results are not available for the KITTI \textit{test} set.

In addition, we also report the performance of the most important ``Car'' class on the $val$ set, together with the runtime of each framework,  in Table~\ref{tab:val_r11}. Here we exploit the average precision of recall 11 positions calculation, so that we can make fair comparison with previous methods. Since the runtime of different methods are obtained with difference devices in their original papers, it is unfair to directly compare them. Therefore, we test these methods using our device with RTX 2080Ti GPU, using the official released codes and models. Both the FPS reported in the corresponding papers (on the left side of `/') and that with our device (on the right side of `/') are included.  Among the methods, our $\text{H}^2$3D R-CNN achieves the best average precision for the moderate and difficult levels on the $\textit{val}$ set. Remarkably, our method outperforms the most related approach, \emph{i.e.}, MVF, by 6.08\% and 2.65\% AP on these two levels.

\begin{table}[!t]
	\centering
	\caption{Performance comparison on the KITTI \textit{val} set in terms of average precision (AP) calculated with recall 11 positions. Both the FPS reported in the corresponding papers (on the left side of `/') and that with our device (on the right side of `/') are included. Here, we add  different symbols after the name of each method to indicate the device used in the original papers: $\dag$--GTX 1080Ti, $\ddag$--RTX 2080Ti, $\S$--Titan V, $\P$--GTX 1070.
	}
	\renewcommand\arraystretch{1.1}
	\footnotesize
	\begin{center}
		\scalebox{0.99}[0.99]{
			\setlength\tabcolsep{7.2pt}
			\begin{tabular}{l|ccc|c}
				\hline
				\multirow{2}{*}{Method} & 				
				\multicolumn{3}{c|}{$\text{Car-AP}_\text{3D}$ (\%) ~~} &
				FPS \\			
				&Easy & Moderate & Hard & (Hz)\\
				\hline
				\hline
				\textbf{3D Feature:} &&& &\\
				SECOND \cite{yan2018second}~$\dag$ & 88.61 & 78.62 & 77.22 & 20.0~/~30.4\\
				PointRCNN \cite{shi2019pointrcnn}  & 88.88 &78.63& 77.38  & ~~~~-~/~10.0\\

				Part-$A^2$ \cite{shi2020points} & 89.47 & 79.47 & 78.54 & ~~~~-~/~10.1\\
				3DSSD \cite{Yang_2020_CVPR}~$\S$  &89.71 & 79.45 & 78.67& 26.3~/~29.1\\
				SA-SSD \cite{He_2020_CVPR} $\ddag$ &  90.15 & 79.91 & 78.78& 25.0~/~23.7\\
				PointGNN \cite{Shi_2020_CVPR_PointGNN}~$\P$& 87.89 & 78.34 & 77.38 & 1.6~/~-~~~~\\
				PV-RCNN \cite{Shi_2020_CVPR} &  89.35 & 83.69 & 78.70 & ~~-~/~8.9~\\
				Voxel R-CNN \cite{deng2020voxelrcnn}~$\ddag$ & 89.41 & 84.52 & 78.93 & 25.2~/~24.8\\
				\hline
				\textbf{BEV Feature:} &&&&\\
				PointPillars \cite{lang2019pointpillars}~$\dag$  &86.62 &76.06 &68.91  & 42.4~/~43.7\\
				TANet \cite{liu2020tanet} ~$\S$ & 87.52 & 76.64 & 73.86 & 28.7~/~31.2\\
				HVNet \cite{Ye_2020_CVPR}~$\ddag$  & 87.21 & 77.58 & 71.79 & 31.0~/~-~~~~~\\
				
				\hline
				\textbf{Multi-View Feature:} &&& \\
				MVF~\cite{zhou2020end}  & \textbf{90.23} & 79.12 & 76.43 & ~-~/~-~\\
				$\text{H}^2$3D R-CNN (ours)  & 89.63 & \textbf{85.20} & \textbf{79.08} & 37.1\\
				\hline
			\end{tabular}
		}
	\end{center}
	
	\label{tab:val_r11}
\end{table}

	\begin{table*}[!t]
		\centering
		\caption{Performance comparison on the Waymo Open Dataset with 202 validation sequences ($\sim$40k samples) for vehicle detection in LEVEL\_1. LEVEL\_1 indicates there are at least 5 points in each instance.
		}
		\renewcommand\arraystretch{1.12}
		\footnotesize
		\begin{center}
			\scalebox{1.0}[1.0]{
				\setlength\tabcolsep{12pt}
				\begin{tabular}{l|cccc|cccc}
					\hline
					\multirow{2}{*}{Method} &
					\multicolumn{4}{c|}{$\text{LEVEL\_1 Vehicle-AP}_\text{3D}$ (\%) ~~} &
					\multicolumn{4}{c}{$\text{LEVEL\_1 Vehicle-AP}_\text{BEV}$ (\%) ~~}\\
					&Overall & 0-30m & 30-50m & 50-Inf 	&Overall & 0-30m & 30-50m & 50-Inf \\
					\hline \hline
					\textbf{3D Feature:} & & & & & & & \\
					CVCNet~\cite{chen2020every} & 65.20 & 86.80 & 63.84 & 36.65 &- & - & - &-\\
					PV-RCNN~\cite{Shi_2020_CVPR}& {70.30} & {91.92} & {69.21} & {42.17} & {82.96} & {97.35} & {82.99} & {64.97} \\ 
					\hline
					\textbf{BEV Feature:} & & & & & & & \\
					PointPillars~\cite{lang2019pointpillars} & 56.62 & 81.01 & 51.75 & 27.94  & 75.57 & 92.1 & 74.06 & 55.47 \\
					AFDet-PP-0.10~\cite{ge2020afdet} & 63.69 & 87.38 & 62.19 & 29.27 & - & - & - &-\\
					
					\hline
					\textbf{Multi-View Feature:} & & & & & & & \\
					MVF~\cite{zhou2020end} & 62.93 & 86.30 & 60.02 & 36.02 & 80.40 & 93.59 & 79.21 & 63.09 \\
					Pillar-od~\cite{wang2020pillar} & 69.80 & 88.53& 66.50& 42.93 & 87.11 & 95.78 & 84.74 & 72.12 \\
					\textbf{$\text{H}^2$3D R-CNN (ours)} & \textbf{75.15} & \textbf{91.98} & \textbf{73.68} & \textbf{52.93} & \textbf{87.84} & \textbf{97.48} & \textbf{87.10} & \textbf{75.47} \\
					\hline
				\end{tabular}
			}
		\end{center}
		
		\label{tab:waymo_veh}
	\end{table*}

	\begin{table}[!t]
		\centering
		\caption{Performance of our $\text{H}^2$3D R-CNN on the Waymo Open Dataset for vehicle detection in LEVEL\_2. LEVEL\_2 includes instances with at least 1 point.
		}
		\renewcommand\arraystretch{1.3}
		\footnotesize
		\begin{center}
			\scalebox{0.98}[0.98]{
				\setlength\tabcolsep{4pt}
				\begin{tabular}{cccc|cccc}
					\hline
					\multicolumn{4}{c|}{$\text{LEVEL\_2 Veh.-AP}_\text{3D}$ (\%) ~~} &
					\multicolumn{4}{c}{$\text{LEVEL\_2 Veh.-AP}_\text{BEV}$ (\%) ~~}\\
					Overall & 0-30m & 30-50m & 50-Inf 	&Overall & 0-30m & 30-50m & 50-Inf \\
					\hline
					\hline
					66.14 & 91.25 & 65.94 & 40.60  & 80.90& 94.55 & 81.27 & 62.73 \\
					\hline
				\end{tabular}
			}
		\end{center}
		
		\label{tab:waymo_level2}
	\end{table}

	\subsubsection{Waymo Open Dataset}
	To further validate the effectiveness of our $\text{H}^2$3D R-CNN, we compare our method with several top-performing approaches on the challenging Waymo Open Dataset. Table~\ref{tab:waymo_veh} shows the results under LEVEL\_1 difficulty in terms of 3D AP and BEV AP. 	
	
	As discussed in Section~\ref{sec:compare_to_mvf}, the multi-view baseline method MVF projects the 3D points into perspective view and the bird-eye view for feature extraction in parallel, and leverages the multi-view features to augment the input of PointPillars~\cite{lang2019pointpillars}. Typically, the over-squeezed height information caused by detecting from the bird-eye view, together with the imbalance issue stemming from using anchors, lead to the sub-optimal performance of MVF, which is 62.93\% AP on this dataset. Pillar-od introduces the anchor-free paradigm~\cite{tian2019fcos} to the MVF framework, partially addressing the imbalance issue of anchors. Thus, the performance boosts to 69.80\%. However, both MVF and Pillar-od perform detection from the bird-eye view, which losses the 3D structure information of point clouds. In the contrast, By taking advantage of the hollow-3D property of point clouds, our $\text{H}^2$3D R-CNN hallucinates powerful 3D representations with the multi-view features, and builds a two-stage framework that refine region proposals in the 3D space. On the one hand, the hallucinated H3D features can better preserve the 3D structure context. On the other hand, the two-stage paradigm with region proposal generation and refinement can also address the problem of imbalanced positive and negative samples at the training stage.

	Our method achieves 75.15\% 3D AP in the overall range by only exploiting 1/5 training data.  
	Compared to the strongest competitor PV-RCNN~\cite{Shi_2020_CVPR}, our method achieves  4.85\% absolute improvement for the overall 3D AP, which further validates that our proposed hallucinated hollow-3d feature is capable  to effectively capture and preserve the accurate contextual information for 3D object detection. Moreover, since we only exploit 2D CNNs in our sequential multi-view feature extraction module, it is much more efficient to generate the hollow-3D features than the point-voxel 3D features. Specifically, the PV-RCNN runs at less than $2$ FPS on an NVIDIA RTX 2080Ti GPU to process a single scene from Waymo Open Dataset, while our proposed $\text{H}^2$3D R-CNN  can run at  $13.6$ FPS on the same device. 
	Since few methods have reported the performance on LEVEL\_2 difficulty, it is difficult to provide comprehensive comparison in this level. To facilitate the comparison of methods for the future, we additionally provide our results on LEVEL\_2 split in Table~\ref{tab:waymo_level2}.

	\subsection{Ablative Experiments}
	We conduct ablative experiments on the KITTI $val$ set~\cite{geigerwe} to extensively verify the effectiveness of our framework.

	\subsubsection{Sequential Multi-View Feature Extraction.}
	In Table~\ref{tab:ablation_mv}, we analyze the effects of the bird-eye view and perspective view in our sequential multi-view feature extraction module. When we only use the bird-eye view for feature extraction and make prediction with the RPN in the bird-eye view (method (a) in the table), the model degenerates to PointPillars~\cite{lang2019pointpillars} (different in the number of filters in the 2D backbone) and achieves a relatively low accuracy with 77.13\% moderate 3D AP. By integrating the perspective-view features, the performance of method (b) boosts to 78.62\%, which validates the semantic-aware perspective-view  feature can ameliorate the bird-eye view features. The method (c) adds a detect head on the top of method (a), and extracts RoI features with HV RoI pooling from the point-wise BEV features $f^{bev}$. As shown in this table, there is a large margin between (c) and our overall model (d), which is due to the absence of perspective-view features in both generating region proposals and box refinement. This result further verifies the significance of sequential multi-view feature extraction for reconstructing the hollow-3D features and facilitating 3D object detection.

		\begin{table}
	\centering
	\caption{Analysis of the bird-eye view and perspective view in our sequential multi-view feature extraction module. Notation: PV indicates the perspective view and BEV indicates the bird-eye view. The moderate 3D AP reported in this table is calculated with the recall 11 positions. 
	}
	
	\renewcommand\arraystretch{1.3}
	\footnotesize
	\begin{center}
		\scalebox{0.93}[0.93]{
			\setlength\tabcolsep{4pt}
			\begin{tabular}{c|cccc|c|c}
				\hline
				Method & BEV & PV  & RPN  & Detect Head & Mod. $\text{AP}_\text{3D}$ (\%)& Runtime (ms) \\
				\hline  \hline
				(a) & $\checkmark$ &  &   $\checkmark$&    & 77.13  & 12.2 \\
				(b) & $\checkmark$ &  $\checkmark$   &  $\checkmark$ &   & 78.62  & 16.8 \\
				(c) &$\checkmark$ &    & $\checkmark$  & $\checkmark$ & 79.33  &  24.2\\
				(d) &$\checkmark$ & $\checkmark$ & $\checkmark$ & $\checkmark$ & 85.20 & 26.9  \\
				\hline
			\end{tabular}
		}
	\end{center}
	
	\label{tab:ablation_mv}
\end{table}

	\begin{figure*}[t]
	\centering {\includegraphics[width=0.97\textwidth]{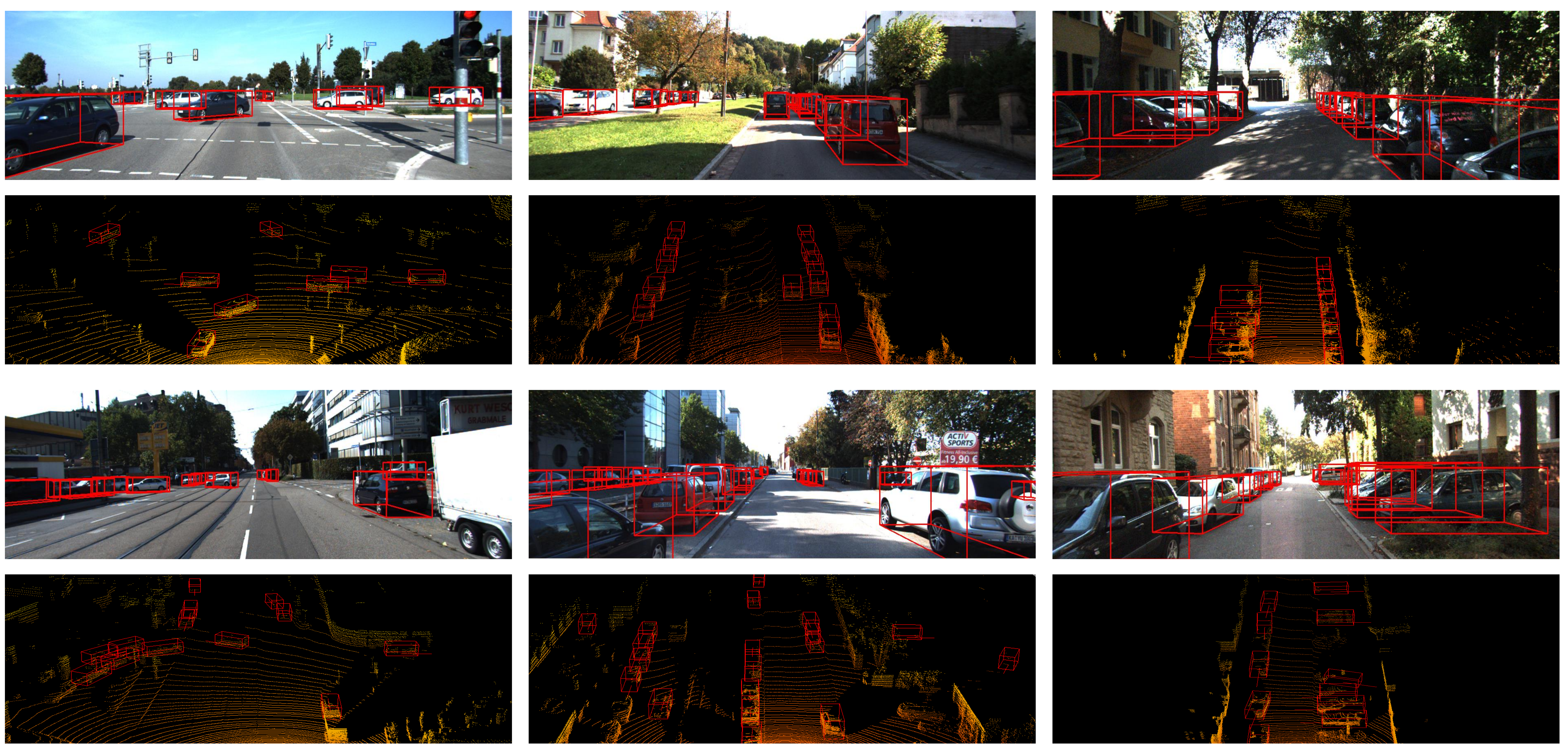}}
	\caption{ Qualitative results on the \textit{test} set of KITTI Dataset.  We showcases six complex scenes with both RGB and LiDAR points in the camera FOV.
	}
	\label{fig:example}
\end{figure*}

\begin{table}
	\centering
	\caption{Analysis of the multi-view features and fusion operation in the hollow-3D feature reconstruction module. Notation: PV indicates the perspective view and BEV indicates the bird-eye view. The moderate 3D AP reported in this table is calculated with the recall 11 positions. 
	}
	\renewcommand\arraystretch{1.3}
	\footnotesize
	\begin{center}
		\scalebox{0.92}[0.92]{
			\setlength\tabcolsep{5.5pt}
			\begin{tabular}{c|cc|ccc|c}
				\hline
				\multirow{2}{*} {Method}
				&\multicolumn{2}{c|}{Feature Source} 
				&\multicolumn{3}{c|}{Fusion Operation}
				& \multirow{2}{*} {Mod. $\text{AP}_\text{3D}$ (\%)} \\ 
				
				&BEV & PV  & Sum  & Concat. & BGMVF& \\
				\hline \hline
				(a)&$\checkmark$ &  &   &    & & 79.67 \\
				(b)& &  $\checkmark$ &   &    & & 83.07 \\
				(c) & $\checkmark$ &  $\checkmark$ &  $\checkmark$  &    & & 83.55 \\
				(d) &$\checkmark$ &  $\checkmark$ &    &   $\checkmark$ & &  84.23 \\
				(e) & $\checkmark$ &  $\checkmark$ &    &   & $\checkmark$  & 85.20  \\
				\hline
			\end{tabular}
		}
	\end{center}
	
	\label{tab:ablation_h3d}
\end{table}

	\subsubsection{Hollow-3D Feature Reconstruction.}
	To further demonstrate the effectiveness of multi-view features and our proposed bilaterally guided multi-view fusion layer, we conduct analytical experiments on the hollow-3D feature reconstruction module. As shown in Table~\ref{tab:ablation_h3d}, when we only utilize the features from the bird-eye view (method (a)) or the features from the perspective-view (method (b)) to reconstruct the hollow-3D features, the performance is much worse than our proposed multi-view-based paradigm (method (e)). Particularly, there is no fusion operator in methods (a) and (b).This verifies the significance of each view in hollow-3D feature reconstruction.

	Besides, from the last three rows in this table, we can observe that the reconstructed hollow-3D features with simple combination methods like element-wise sum or concatenation also lead to sub-optimal performance. The main reason is that there is a large information gap between our multi-view features, \emph{i.e.}, semantic-aware for the perspective view and position-aware for the bird-eye view, while the simple combination fails to make full use of them. Typically, our bilaterally guided multi-view fusion (BGMVF) layer exploits the cross-view attention to guide feature selection before combination, which benefits the interaction of multi-view features. Capitalizing on BGMVF layer, our $\text{H}^2$3D R-CNN achieves reasonable performance gain compared to that with simple combinations, \emph{i.e.}, 1.65\% over element-wise sum and 0.97\% over concatenation. Here we ignore the runtime comparison in this table, since the cost of changing the element-wise sum or concatenation to our BGMVF is negligible compared to the whole framework. 
	
		\begin{table}
	\centering
	\caption{Ablative experiments of the coarse grids partition and fine grids partition in our HV RoI Pooling. Here we set the coarse partition as 3$\times$3$\times$3 and set the fine partition as 6$\times$6$\times$6. The moderate 3D AP reported in this table is calculated with the recall 40 positions. 
	}
	\renewcommand\arraystretch{1.3}
	\footnotesize
	\begin{center}
		\scalebox{0.95}[0.95]{
			\setlength\tabcolsep{5pt}
			\begin{tabular}{c|cc|c|c}
				\hline
				Method & Coarse Grids& Fine Grids & Mod. $\text{AP}_\text{3D}$ (\%) & Runtime (ms) \\
				\hline \hline
				(a) &$\checkmark$ &  &   83.10 & 21.8 \\
				(b) & &  $\checkmark$ &   84.29 & 24.0\\
				(c) & $\checkmark$ &  $\checkmark$ &  85.62 & 26.9 \\
				\hline
			\end{tabular}
		}
	\end{center}
	
	\label{tab:ablation_hv_roi}
\end{table}

	\subsubsection{Hierarchical Voxel RoI Pooling.} Table~\ref{tab:ablation_hv_roi} shows the performance of adopting coarse, fine and hierarchical grids partition. As shown in the table, when only using coarse grids, \emph{i.e.}, 3$\times$3$\times$3 grids partition, the model runs fast, but with an unsatisfied 83.10\% 3D AP. By replacing the coarse grids with fine grids, \emph{i.e.}, 6$\times$6$\times$6 grids partition, the performance boosts to 84.29\% 3D AP. Our proposed HV RoI Pooling, instead, introduces hierarchical partition when performing RoI feature extraction, which achieves 1.33\% performance gain with only 2.9ms processing latency. As mentioned in the implementation details, our HV RoI Pooling only utilizes a single query radius for grids belonging to each granularity. In particular, another design choice is to exploit the fine grid partition with multi-scale query radii, which degenerates to the conventional Voxel RoI Pooling in~\cite{deng2020voxelrcnn}. In this setting, the average precision at the moderate level is 84.71\%, 0.91\% lower than our overall model (d), and the runtime cost is 34.3ms, 7.4ms slower than our overall model (d). The results validate our proposed HV RoI Pooling to exploit hierarchical grids partition steadily leads to improvements in both accuracy and efficiency. 
	
	\subsection{Qualitative Results.} 
	We showcase the qualitative results of several scenes from the KITTI $test$ set in Figure \ref{fig:example}. As shown in the figure, the proposed $\text{H}^2$3D R-CNN can precisely detect the objects even on the challenging crossroad or  in a single scene with more  than 17 cars. These qualitative examples further provide intuitive evidence on the effectiveness of our approach.

	\section{Conclusion}
	In this work, we are dedicated to 3D object detection and propose a new Hallucinated Hollow-3D R-CNN framework. To take the advantage of the hollow 3D property of point clouds, we explore the complementary information in the  perspective view and the bird-eye view, and hallucinates from the above two views into 3D  representation by a novel bilaterally guided  multi-view  fusion block. Moreover, the RoI representation is extracted from the hollow-3D features with HV RoI Pooling for further acceleration. Our design makes the best of hollow 3D representation and 
	establishes a careful trade-off between precision and runtime latency. Extensive experiments with qualitative results on two benchmark datasets, together with quantitative examples, demonstrate the effectiveness and efficiency of our proposed method.

	
	\bibliographystyle{IEEEtran}
	\bibliography{./reference}

\end{document}